\documentclass[letterpaper]{article} 
\usepackage{aaai2026}  
\usepackage{times}  
\usepackage{helvet}  
\usepackage{courier}  
\usepackage[hyphens]{url}  
\usepackage{graphicx} 
\usepackage{natbib}  
\usepackage{caption} 
\usepackage{algorithm}
\usepackage{algorithmic}

\usepackage{newfloat}
\usepackage{listings}
\usepackage{siunitx}
\DeclareCaptionStyle{ruled}{labelfont=normalfont,labelsep=colon,strut=off} 
\floatstyle{ruled}
\newfloat{listing}{tb}{lst}{}
\floatname{listing}{Listing}
\usepackage{multirow, makecell}
\usepackage{array}
\usepackage{cutwin}
\usepackage{marvosym}
\usepackage{booktabs}
\usepackage{amsmath, amssymb}
\usepackage{pifont}

\newcolumntype{L}[1]{>{\raggedright\arraybackslash}p{#1}}   
\newcolumntype{C}[1]{>{\centering\arraybackslash}p{#1}}     
\newcolumntype{R}[1]{>{\raggedleft\arraybackslash}p{#1}}    

\title{DuGI-MAE: Improving Infrared Mask Autoencoders via Dual-Domain Guidance}
\author{
    Yinghui Xing\textsuperscript{\rm 1}~~Xiaoting Su\textsuperscript{\rm 1}~~Shizhou Zhang\textsuperscript{\rm 1}\thanks{Corresponding Author:szzhang@nwpu.edu.cn}~~Donghao Chu\textsuperscript{\rm 1}~~Di Xu\textsuperscript{\rm 2}
}
\affiliations{
    \textsuperscript{\rm 1}School of Computer Science, Northwestern Polytechnical University, China ~~~~
    
    \textsuperscript{\rm 2}	Huawei, China
}

\begin{document}

\maketitle

\begin{abstract}
Infrared imaging plays a critical role in low-light and adverse weather conditions. However, due to the distinct characteristics of infrared images, existing foundation models such as Masked Autoencoder (MAE) trained on visible data perform suboptimal in infrared image interpretation tasks. To bridge this gap, an infrared foundation model known as InfMAE~\cite{liu2024infmae} was developed and pre-trained on large-scale infrared datasets. Despite its effectiveness, InfMAE still faces several limitations, including the omission of informative tokens, insufficient modeling of global associations, and neglect of non-uniform noise. In this paper, we propose a Dual-domain Guided Infrared foundation model based on MAE (DuGI-MAE). First, we design a deterministic masking strategy based on token entropy, preserving only high-entropy tokens for reconstruction to enhance informativeness. Next, we introduce a Dual-Domain Guidance (DDG) module, which simultaneously captures global token relationships and adaptively filters non-uniform background noise commonly present in infrared imagery. To facilitate large-scale pretraining, we construct Inf-590K, a comprehensive infrared image dataset encompassing diverse scenes, various target types, and multiple spatial resolutions. Pretrained on Inf-590K, DuGI-MAE demonstrates strong generalization capabilities across various downstream tasks, including infrared object detection, semantic segmentation, and small target detection. Experimental results validate the superiority of the proposed method over both supervised and self-supervised comparison methods.
\end{abstract}


\section{Introduction}
Infrared (IR) imaging has emerged as a critical sensing modality in various applications, including surveillance~\cite{jia2021llvip} and target detection~\cite{xing2024ms}.
Compared to visible images, infrared images exhibit unique characteristics such as lower spatial detail, reduced texture information, and generally lower signal-to-noise ratios. Their content is primarily governed by thermal radiation, making them highly sensitive to temperature differences but also less informative in scenes with low thermal contrast, all of which pose significant challenges for accurate visual interpretation.

\begin{figure}[t]
    \centering
    \includegraphics[width=\columnwidth]{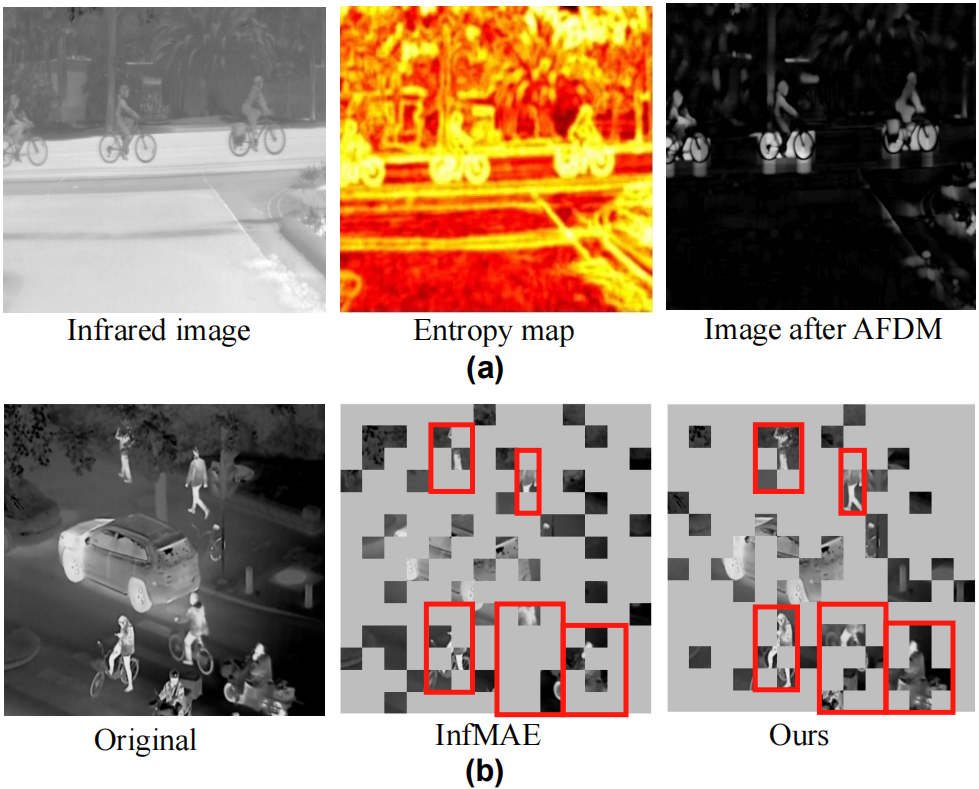}
    \caption{
    (a) Representative infrared image from a typical scene. Left: The original infrared image, where strong background responses often suppress the actual targets; Middle: Entropy map of the image; Right: Image processed with Adaptive Frequency-Domain Modulation (AFDM).
    (b) Comparison between Information-aware masking~\cite{liu2024infmae} and our Entropy-based masking.}
    \label{fig:motivation}
\end{figure}

As illustrated in Figure~\ref{fig:motivation}, infrared images lack texture, and the temperature difference between the background and the target can cause the target to be suppressed, as in Figure~\ref{fig:motivation}(a), or highlighted, as in Figure~\ref{fig:motivation}(b).
Therefore, models trained on visible images cannot be directly transferable, emphasizing the necessity of designing infrared modality-specific learning frameworks. 
Recent advances in modality-specific methods for infrared imagery include YOLO-Infrared~\cite{zhang2022yolo} and PFGF~\cite{li2025pseudo} for object detection in natural scenes, TBC-Net~\cite{zhao2019tbc} for semantic segmentation, and IRSTD~\cite{liu2024infrared} and SCAFNet~\cite{zhang2024scafnet} for small target detection.
These methods learn task-specific representations through supervised training on dedicated datasets. However, due to the inherent modality gap between visible and infrared imagery, transferring features from ImageNet often results in suboptimal performance.

Vision foundation models have demonstrated remarkable generalization capabilities across a wide range of tasks, largely attributed to self-supervised pre-training on large-scale image datasets.
Among them, the Masked Autoencoder (MAE)~\cite{he2022masked} adopts masked reconstruction as a pretext task to learn transferable representations by reconstructing the masked tokens. MAE has achieved remarkable results in various downstream vision tasks. However, its random masking strategy is ineffective in infrared modality, since infrared images inherently have low information density, randomly masking out the informative tokens leads difficulties in reconstruction. 
Recently,~\cite{liu2024infmae} proposed InfMAE, an infrared foundation model that employs an information-aware masking strategy to evaluate the information richness of each region based on its gray value.
Although this strategy considers the characteristics of infrared images, it has the following limitations: \textbf{1) Omission of informative tokens.} InfMAE utilizes a sampling strategy, which samples tokens with a fixed interval, resulting in the possible omission of some informative tokens. \textbf{2) Insufficient modeling of global association.} InfMAE lacks a global association mechanism; if the tokens used for reconstruction are spatially dispersed within the image, InfMAE is ineffective in reconstructing the masked tokens. \textbf{3) Neglect of non-uniform noise.} 
Due to the factors such as variations in detector responsivity, thermal instability, and imperfections in the optical system~\cite{fang2025detection}, infrared images often suffer from non-uniform noise. Figure~\ref{fig:motivation}(a) illustrates a type of non-uniform noise, temperature drift noise, which tends to suppress target regions while amplifying background responses. In this case, the gray-value-based masking strategy fails to account for non-uniform noise, leading to background regions being mistakenly retained as informative tokens.

To address the aforementioned problems, we propose a \textbf{Du}al-domain \textbf{G}uided \textbf{I}nfrared foundation model based on \textbf{MAE}, termed \textbf{DuGI-MAE}. Specifically, we first design a deterministic masking strategy that selectively retains the most informative tokens. This ``non-sampling'' masking fundamentally avoids the loss of critical information often caused by random or fixed-interval sampling. To enhance global association and simultaneously suppress non-uniform noise, we further introduce a Dual-Domain Guidance (DDG) module, which incorporates an adaptive frequency filter. The integration of frequency-domain features is motivated by prior studies~\cite{he2023fspnp, shi2024semi}, which demonstrate their effectiveness in capturing global spatial structures and attenuating non-uniform noise in infrared images.
The DDG module serves as a bridge between the encoder and decoder, enhancing the learning of robust and noise-resistant infrared representations. 
To facilitate pre-training, we construct a large-scale infrared dataset, Inf-590K, comprising 590,700 infrared images collected from diverse platforms and viewpoints, encompassing a wide range of scenes, target types, and spatial scales. Leveraging this large-scale pre-training, DuGI-MAE substantially outperforms state-of-the-art methods across various downstream tasks.

The main contributions are summarized as follows:  
\begin{itemize}
    \item We propose a dual-domain guided foundation model, DuGI-MAE, which uses a deterministic entropy-based masking strategy to mitigate missing informative tokens.  
    \item We present the DDG module to guide masked token reconstruction, employing adaptive frequency filtering to reduce non-uniform noise in infrared images. 
    \item We construct a large-scale dataset, \textbf{Inf-590K}, specifically for self-supervised pretraining on infrared imagery. Pretraining on Inf-590K significantly improves the generalization ability of various self-supervised methods for infrared image interpretation tasks.
    \item Experimental results on infrared object detection, semantic segmentation, and small target detection consistently show the superiority and generalizability of DuGI-MAE.  
\end{itemize}

\section{Related Work}
\textbf{Vision Foundation Model.}
Vision foundation models, which learn general image representations through large-scale self-supervised pre-training, have emerged as a dominant paradigm in computer vision. Most existing foundation models have primarily focused on visible images, with representative approaches including Masked Autoencoders (MAE) \cite{he2022masked}, Bidirectional Encoder Representation from Image Transformers (BEiT) \cite{bao2021beit}, and Self-Distillation with No Labels (DINO) \cite{caron2021emerging}. These models are typically pre-trained on large-scale datasets such as ImageNet, leading to substantial performance gains in various downstream tasks.
However, they are inherently designed under the assumption that images contain rich texture and color information. Such assumptions present fundamental limitations when applied to infrared imagery, which typically lacks detailed textures and chromatic information. Recently, \cite{liu2024infmae} proposed InfMAE for the infrared modality, which used an Information-Aware Masking strategy to retain the informative regions by measuring their gray values. They further designed a multi-scale encoder to enhance local feature learning, achieving significant progress in infrared modality related tasks. 
InfMAE is pre-trained on a dataset of 300K samples, and its performance cannot be further improved without introducing more diverse data.
Additionally, challenges still persist in the masking strategy and encoding process.

\textbf{Frequency Domain in Infrared Image Processing.}
As a fundamental technique in signal processing~\cite{pitas2000digital}, frequency-domain analysis enables powerful modeling of the distinct thermal radiation properties inherent in infrared imagery~\cite{2012Infrared,2011Infrared}.
Recent advances in infrared image processing have increasingly leveraged frequency-domain information to enhance feature representation and discrimination. 
For example, \cite{10214008} introduced Fourier transform layers into convolutional neural network (CNN) backbones, enabling the model to learn spectral features that better characterize infrared targets. 
Similarly, \cite{2024Dual} proposed a dual-domain fusion framework that integrates spatial and frequency-domain features at multiple levels, effectively capturing both local thermal edges and global scene structures, and demonstrating superior performance in infrared semantic segmentation.
In contrast to these approaches, our work emphasizes the frequency-domain in the self-supervised pretext task to learn generalizable feature representations.

\section{Infrared Pre-training Dataset: Inf-590K}
\begin{figure}[t]
    \centering
    \includegraphics[width=\columnwidth]{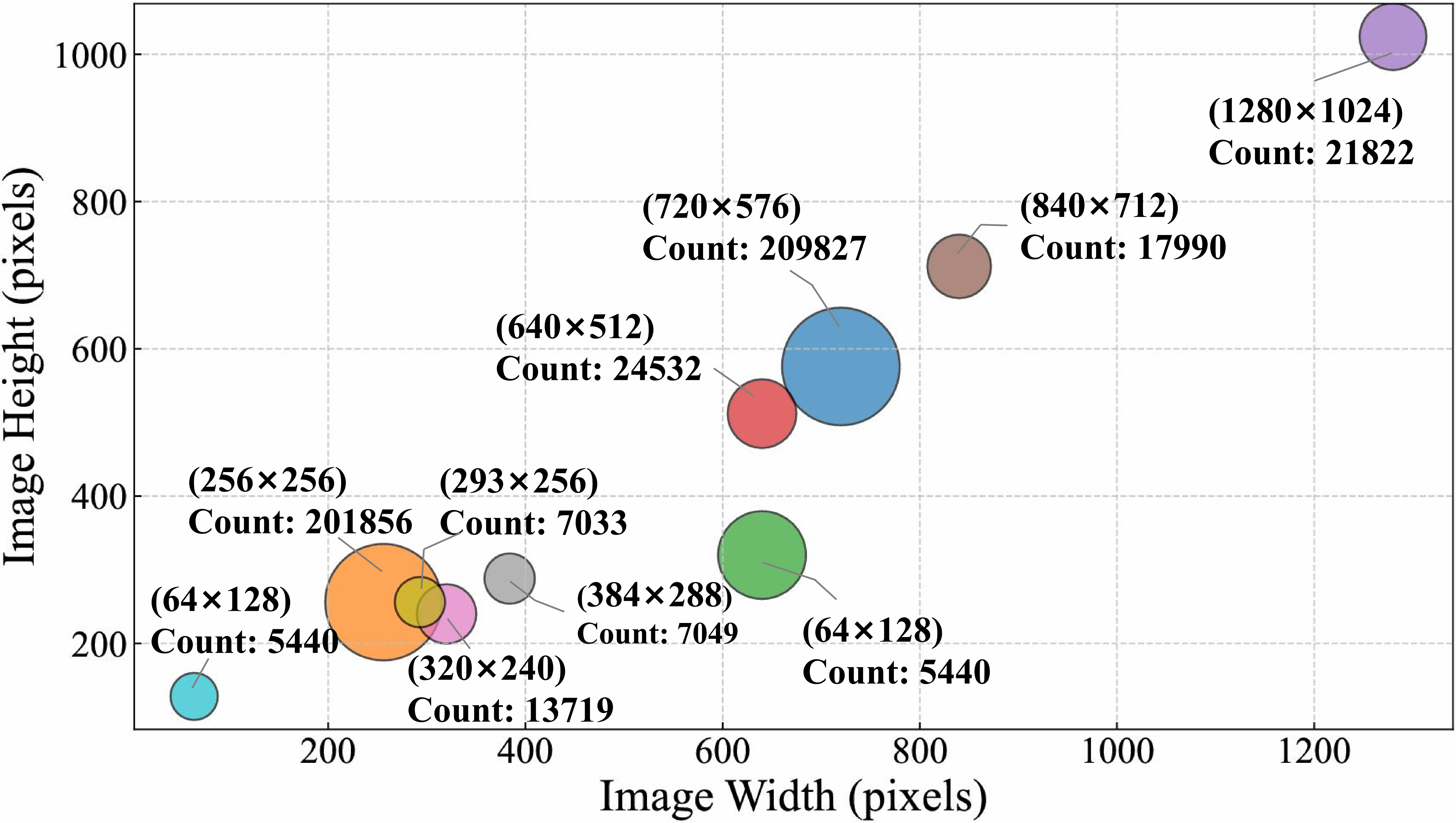}
    \caption{Resolution distribution of the Inf-590K dataset. The horizontal and vertical axes represent image width and height, respectively, while the size of each bubble indicates the number of samples corresponding to that resolution. }    \label{fig:dataset}
\end{figure}

We construct a large-scale infrared pre-training dataset named Inf-590K, comprising 590,700 infrared images. The primary data sources include several publicly available infrared datasets~\cite{jia2021llvip,2022Drone,2022Visible,2013Multispectral,2020FusionDN,liu2020lsotb,liu2024infmae} as well as self-collected real-world infrared images. 
The dataset covers a variety of acquisition platforms and viewpoints, including aerial surveillance perspectives captured by UAV platforms, monitoring views from ground-based fixed cameras, street-level perspectives acquired by vehicle-mounted sensors, and maritime monitoring views obtained from ship-borne sensors.
The diverse acquisition conditions contribute to a variety of scenarios and perspectives, but may also lead to a certain degree of data redundancy.
To eliminate the redundant samples, we randomly select an anchor image from each scene and compute the cosine similarity between the features of this anchor and those of candidate images. 
Samples with high similarity scores (i.e., above 0.85) are excluded, thereby retaining images that exhibit distinct scene layouts or target distributions. 
Additionally, for visible-infrared video pairs, misalignment between imaging sensors often results in black borders in the infrared frames. To address this issue, we identify border regions with zero pixel values and apply adaptive cropping to remove these invalid areas.

After preprocessing, the Inf-590K dataset includes 445 unique resolutions ranging from a minimum of 50$\times$34 to a maximum of 6912$\times$576, showcasing a wide range of spatial resolutions.
Figure~\ref{fig:dataset} presents the top 10 most common image resolutions. 
This distribution aligns with the typical resolution characteristics of infrared imaging systems used in practical applications.
In terms of scene diversity, Inf-590K comprises terrestrial (e.g., urban areas, highways, rural terrain), maritime (e.g., coastlines, open sea), and aerial (e.g., clouds, aerial views of ground targets) scenes, encompassing a variety of weather conditions (clear, foggy, rainy) and temporal scenarios (day and night). The annotated targets span common infrared objects, including vehicles, pedestrians, buildings, ships, aircraft, and public infrastructure.

\section{Proposed Method}
\begin{figure*}[!t]
    \centering
    \includegraphics[width=1.0\linewidth]{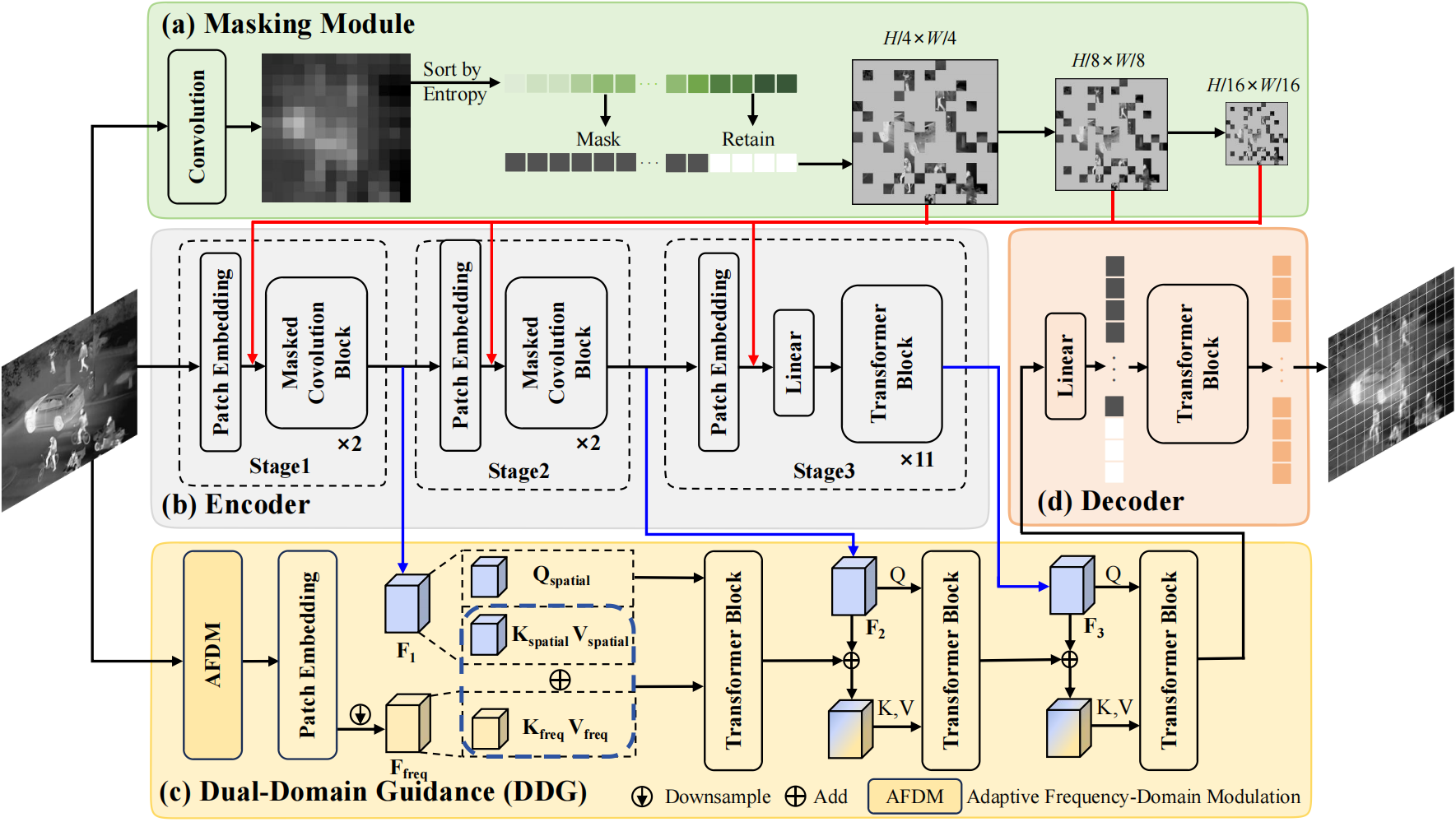}
    \caption{Overall architecture of DuGI-MAE. It consists of the (a) Entropy-Based Masking Module, the (b) Encoder, the (c) Dual-Domain Guidance (DDG) module, and the (d) Decoder.}
    \label{fig:overallframework}
\end{figure*}

\subsection{Overview}
The overall framework of proposed DuGI-MAE is shown in~Figure~\ref{fig:overallframework}.
The pretext task for self-supervised learning (SSL) is the masked image modeling (MIM). Driven by our entropy-based masking strategy, the masked autoencoder (MAE)~\cite{he2022masked} reconstructs the masked tokens to capture inherent thermal radiation characteristics. Meanwhile, we propose a dual-domain guidance (DDG) module, where frequency features act as a condition to fuse spatial structural and thermal radiation information, thereby guiding the decoder to reconstruct masked tokens.

\subsection{Entropy-Based Masking Module}
To make masked image reconstruction a meaningful pretext task, previous studies have commonly applied aggressive masking by randomly masking a substantial portion of input image tokens~\cite{he2022masked,wei2022masked}. This strategy, however, leads to the difficulties in reconstructing foreground details~\cite{hou2022milan} if the remaining visible tokens comprise more background information. 

For the infrared modality, precise selection of masked positions is particularly important, as infrared images often exhibit sparse thermal radiation distributions. In this paper, we employ \textbf{Shannon Entropy} as a quantitative metric to assess the information content of each token. Tokens with higher entropy values are prioritized for retention, while a higher proportion of masking is applied to other regions. This strategy encourages the model concentrate on discriminative features specific to the infrared modality. 
As shown in Figure~\ref{fig:motivation}(b), the retained tokens consistently preserve informative content, including target regions and the boundaries between targets and the background.

Given an infrared image $\mathbf{X}$, we first use a convolutional layer $\Phi(\cdot)$ to generate feature map $\mathbf{M} = \Phi(\mathbf{X})$, which is further partitioned and flattened into $N$ tokens $\{\mathbf{m}_1, \mathbf{m}_2, \cdots, \mathbf{m}_N\}$. We then calculate the entropy value of these tokens using the following formula:
\begin{equation}
EN(\mathbf{m}_i) = -\sum_{j=1}^{J} P(h_j) \cdot \log_2\left(P(h_j)\right),
\end{equation}
where $h_j$ denotes the $j$-th intensity value within $\boldsymbol{\mathbf{m}}_i$, $J$ is the total number of possible intensity levels, and $P(h_j)$ represents the probability of $h_j$ occurring in $\boldsymbol{\mathbf{m}}_i$.   

Let $\lambda$ be the mask ratio, we sort the tokens according to their entropy values in an ascending order, and retain the last $(1-\lambda)$ tokens with the highest entropy values, which we think represent the most informative regions in the infrared image, typically containing important thermal targets. 
The remaining tokens with lower entropy are masked, as these regions often indicate uniform backgrounds. This process is formulated as:
\begin{equation}
\begin{aligned}
\hfill &\mathcal{I}_{\text{keep}} = \left\{ \mathcal{I}_{\text{sort}}[i] \mid i \in \left[ \lfloor \lambda \cdot N \rfloor, N - 1 \right] \right\}, \\
\hfill &\mathbf{Mask}[i] = 
\begin{cases} 
1, & \text{if } i \in \mathcal{I}_{\text{keep}}, \\
0, & \text{otherwise},
\end{cases}
\end{aligned}
\end{equation}
where $i$ is the token index, and $\mathcal{I}_{\text{sort}}$ denotes the sorted index sequence. $\mathcal{I}_{\text{keep}}$ represent the tokens to be kept. $\lfloor \cdot \rfloor$ is the floor function. In our method, we set $\lambda$ = 0.75.

\subsection{Dual-Domain Guidance}
Due to the low information density of infrared modality, the preserved tokens used for reconstruction may be dispersed in the image. In this case, the global association is very important. Furthermore, infrared images typically exhibit non-uniform noises, which disrupt the discrimination between targets and background.
Since the frequency domain is capable of modeling global information present in the spatial domain. Additionally, high-pass filtering in the frequency domain can help alleviate non-uniform noise, we incorporate frequency transform into our model. Specifically, we propose a Dual-Domain Guidance (DDG) module that leverages both spatial and frequency domain features to enhance performance.
As illustrated in Figure~\ref{fig:overallframework}(c), frequency features are extracted from infrared images using Adaptive Frequency-Domain Modulation (AFDM). These features are then projected through a patch embedding to serve as key-value pairs for subsequent Transformer block. In parallel, spatial features from the encoder are formulated as query-key-value triplets. The frequency-enhanced features subsequently guide the spatial features, enabling them to more effectively attend to target regions.

\begin{figure}[t]
    \centering
    \includegraphics[width=\columnwidth]{AFDM.png}
    \caption{Adaptive Frequency-Domain Modulation (AFDM). The input images are first transformed into the frequency domain via the Fast Fourier Transform (FFT). A learnable radial filter is then applied to suppress non-uniform background noise(usually low-frequency components) while preserving discriminative features. Finally, the processed features are transformed back to the spatial domain using the Inverse FFT (IFFT).}
    \label{fig:AFDM}
\end{figure}

\textbf{Adaptive Frequency-Domain Modulation.}  
Although the non-uniform noise, especially the temperature drift noise can be alleviated by the high-pass filtering, we argue that directly filtering the low-frequency component leads to the loss of image contents. Therefore, we propose an adaptive frequency-domain modulation (AFDM), shown in Figure~\ref{fig:AFDM}.
Firstly, the Fast Fourier Transform (FFT) converts the input image $\mathbf{X}$ to the frequency domain, obtaining \( \mathcal{F}(\mathbf{X})\), where $\mathcal{F}(\cdot)$ denotes the FFT. In the frequency spectrum, the central region corresponds to low-frequency components.
We design a parameterized radial function 
\(\mathbf{H}(u, v)\) for filtering the frequency spectrum centered at the spectral center, dynamically adjusting central (low-frequency) suppression intensity via learnable parameters while preserving information in the peripheral (mid-to-high frequency) regions:
\begin{equation}
\begin{aligned}
\mathbf{H}(u, v) & = \alpha \cdot \exp\left(-\beta \cdot \left( \frac{D(u, v)}{r} \right)^2 \right), \\
\mathbf{S} & = \mathcal{F}(\mathbf{X}) \odot \mathbf{H},
\end{aligned}
\end{equation}

where $\mathbf{S}$ is the modulated frequency spectrum, and ``$\odot$'' denotes the element-wise multiplication.
$r$, \( \alpha \in [0, 1) \), and \( \beta > 0 \) are learnable parameters, where $r$ is the radius, and $\alpha$ and $\beta$ control the attenuation degree and rate, respectively. $D(u, v)$ is the Euclidean distance from the coordinate \((u, v)\) to the center of the frequency spectrum. By adjusting the suppression parameters \( \alpha \) and $\beta$, the model can dynamically control the attenuation of low-frequency signals.
This frequency modulation selectively attenuates the spectral center to reduce non-uniform noise while preserving high-frequency thermal radiation signals, enhancing the feature discriminability. The modulated frequency spectrum \( \mathbf{S} \) is then transformed back to the spatial domain via Inverse FFT (IFFT), yielding the frequency-enhanced feature map \( \mathbf{F}_{\text{freq}}\) after the patch embedding and downsampling.  

\textbf{Frequency-Guided Attention Injection.} 
We use the frequency-enhanced feature map \( \mathbf{F}_{\text{freq}} \) to act as the guidance for the features in spatial domain.
Specifically, in the first Transformer block, \( \mathbf{F}_{\text{freq}} \) is firstly encoded into Key (\( \mathbf{K}_{\text{freq}} \)) and Value (\( \mathbf{V}_{\text{freq}} \)) pairs, and they are then integrated with the spatial Key (\( \mathbf{K}_{\text{spatial}} \)) and Value (\( \mathbf{V}_{\text{spatial}} \)) pairs:  

\begin{equation}
\begin{aligned}
Atten&tion(\mathbf{F}_{1}, \mathbf{F}_{\text{freq}}) = &\\
&\sigma(\frac{\mathbf{Q}_{\text{spatial}} \left( \mathbf{K}_{\text{spatial}}^T + \mathbf{K}_{\text{freq}}^T \right)}{\sqrt{d_k}}) \cdot (\mathbf{V}_{\text{spatial}} + \mathbf{V}_{\text{freq}}),
\end{aligned}
\tag{5}
\end{equation}
where $\mathbf{F}_1$ represents the output of \textit{Stage 1} in the encoder. $\sigma(\cdot)$ is the softmax function, \( \mathbf{Q}_{\text{spatial}} \) is the query of spatial features, and \( \sqrt{d_k} \) is the scaling factor. 
In the subsequent Transformer block, spatial features are added to the output of the previous block to form the Key and Value representations, while the spatial features themselves continue to serve as the Query.
The frequency-guided attention injection steers attention weights toward high-response target regions to enhance critical structures.  

\subsection{DuGI-MAE for Downstream Tasks}
The pre-trained encoder of DuGI-MAE can be adapted to downstream tasks, including infrared object detection, semantic segmentation, and small target detection.
To enable multi-scale feature aggregation, we extract intermediate features from the encoder, denoted as \( \mathbf{F}_1 \), \(\mathbf{F}_2 \), and \( \mathbf{F}_3 \), which are the outputs of \textit{Stage 1}, \textit{Stage 2}, and \textit{Stage 3}, respectively. These features progressively encode spatial and semantic information at different levels. We additionally apply a downsampling operation to $\mathbf{F}_3 \in \mathbb{R}^{H/16 \times W/16 \times C}$, producing $\mathbf{F}_4 \in \mathbb{R}^{H/32 \times W/32 \times C}$.
Finally, the multi-scale features \( \mathbf{F}_1, \mathbf{F}_2, \mathbf{F}_3, \) and \( \mathbf{F}_4 \) are fed into the downstream tasks for detection and segmentation.

\section{Experiment}
\begin{table}[t]
\centering
\footnotesize  
\setlength{\tabcolsep}{3pt}  
\begin{tabular}{l c c S[table-format=2.1] S[table-format=2.1]}
\toprule
Method & Backbone & Model & {mAP} & {AP$_{50}$} \\
\midrule
DETR & ResNet101 & -- & 41.5 & 72.7 \\
DINO & Swin-L & -- & 44.6 & 74.8 \\
YOLOv8 & CSPDarkNet & -- & 53.7 & 80.3 \\
\midrule
From scratch & VIT-B & Cascade R-CNN & 51.2 & 80.1 \\
MAE & ViT-B & Cascade R-CNN & 51.9 & 82.0 \\
MCMAE & ViT-B & Cascade R-CNN & 55.1 & 85.4 \\
InfMAE & ViT-B & Cascade R-CNN & 56.5 & 86.3 \\
DuGI-MAE(Ours) & ViT-B & Cascade R-CNN & \textbf{57.3} & \textbf{86.9} \\
\midrule
From scratch & VIT-B & Mask R-CNN & 49.6 & 80.3 \\
MAE & ViT-B & Mask R-CNN & 52.2 & 83.8 \\
MCMAE & ViT-B & Mask R-CNN & 56.8 & 87.2 \\
InfMAE & ViT-B & Mask R-CNN & 57.1 & 87.9 \\
DuGI-MAE(Ours) & ViT-B & Mask R-CNN & \textbf{59.1} & \textbf{89.7} \\
\bottomrule
\end{tabular}

\caption{Performance comparisons of different object detection methods on the M$^3$FD-inf dataset.}
\label{tab:detection}
\end{table}

\begin{table}[t]
\centering
\footnotesize
\setlength{\tabcolsep}{4pt}
\begin{tabular}{l c c S[table-format=2.1] S[table-format=2.1]}
\toprule
Method & Backbone & Model & {mIoU} & {mAcc} \\
\midrule
DeeplabV3+ & Resnet50 & -- & 65.2 & 73.8 \\
UperNet & Resnet50 & -- & 65.6 & 74.7 \\
DNLNet & Resnet101 & -- & 67.0 & 75.7 \\
DDRNet & -- & -- & 67.3 & 73.3 \\
\midrule
From scratch & ViT-B & FCN & 57.0 & 63.6 \\
MAE & ViT-B & FCN & 63.4 & 70.8 \\
MCMAE & ViT-B & FCN & 70.8 & 79.2 \\
InfMAE & ViT-B & FCN & 72.6 & 80.5 \\
DuGI-MAE(Ours) & ViT-B & FCN & \textbf{73.1} & \textbf{80.8} \\
\midrule
From scratch & ViT-B & UperNet & 61.5 & 61.3 \\
MAE & ViT-B & UperNet & 71.3 & 78.0 \\
MCMAE & ViT-B & UperNet & 73.2 & 81.1 \\
InfMAE & ViT-B & UperNet & 74.5 & 82.9 \\
DuGI-MAE(Ours) & ViT-B & UperNet & \textbf{75.0} & \textbf{83.6} \\
\bottomrule
\end{tabular}
\caption{Performance comparisons of different semantic segmentation methods on MSRS dataset.}
\label{tab:msrs_segmentation}
\end{table}

\begin{table}[t]
\centering
\footnotesize
\begin{tabular}{l c c c c}
\toprule
Method   & Backbone & Model & mIoU   & Pd     \\
\midrule
MPCM     & --       & --    & 7.3  & 60.3 \\
IPI      & --       & --    & 28.0 & 81.3 \\
RIPT     & --       & --    & 14.1 & 77.5 \\
ACMNet   & --       & --    & 60.3 & 90.3 \\
DNANet   & --       & --    & 65.7 & 89.2 \\
UIUNet   & --       & --    & 65.6 & 91.3 \\
SCAFNet  & --       & --    & 66.3 & 91.1 \\
\midrule
MAE      & ViT-B    & IRSTD  & 57.5   & 88.2   \\
MCMAE    & ViT-B    & IRSTD  & 64.3   & 77.9   \\
InfMAE   & ViT-B    & IRSTD  & 66.5   & \textbf{96.9} \\
DuGI-MAE(Ours)     & ViT-B    & IRSTD  & \textbf{67.1} & 95.9 \\
\bottomrule
\end{tabular}
\caption{Performance comparisons of different infrared small target detection methods on the IRSTD-1k dataset.}
\label{tab:irstd_detection}
\end{table}

In this section, we outline the pre-training setup on Inf-590K dataset and proceed to evaluate our model’s generalization performance through three downstream tasks. Finally, we present comprehensive ablation studies. Our code is available at \url{https://github.com/Xtingsu/DuGI-MAE}.

\subsection{Pre-training Setup}
The DuGI-MAE framework is implemented using PyTorch 1.8.0 and trained on four NVIDIA GeForce RTX 4090 GPUs. In line with the settings commonly adopted in MAE-series self-supervised learning frameworks~\cite{he2022masked,gao2022convmae}, we utilize mean squared error loss during pre-training and employ a mask ratio of 75\%. The encoder is structured into three hierarchical stages comprising 2, 2, and 11 Transformer layers, respectively, to effectively capture multi-scale thermal features.
Pre-training is conducted over 400 epochs using a cosine learning rate schedule, with the initial 40 epochs allocated for warm-up to enhance training stability. The model is optimized with the AdamW optimizer, utilizing a base learning rate of $1.5 \times 10^{-4}$, a weight decay of 0.05, and a batch size of 96. To improve the generalization across diverse infrared scene variations, random cropping is employed as the primary data augmentation strategy. 

\subsection{Infrared Object Detection}
\textbf{Experimental Settings.} 
We validate the effectiveness of proposed method on infrared object detection using the M$^3$FD-inf dataset~\cite{liu2022target}. The M$^3$FD dataset is a multimodal object detection benchmark comprising 4,200 paired infrared and visible images, covering six object categories: person, car, bus, motor, truck, and lamp. In our experiments, only the infrared image subset, referred to as M$^3$FD-inf, is employed for performance evaluation. We adopt Mask R-CNN~\cite{he2017mask} and Cascade R-CNN~\cite{cai2018cascade} as detection heads, with the pre-trained DuGI-MAE model serving as the backbone. The entire network is fine-tuned for 260k iterations using a base learning rate of \(1\times 10^{-6}\) and a weight decay of 0.1. 

\textbf{Results and Analyses.} 
To comprehensively evaluate the effectiveness of the proposed method, we compare it with both fully supervised approaches-DETR~\cite{carion2020end}, DINO~\cite{zhang2022dino}, YOLOv8) and self-supervised methods, including MAE~\cite{he2022masked}, MCMAE~\cite{gao2022convmae}, and InfMAE~\cite{liu2024infmae}. Except for the model trained from scratch, all methods are pre-trained on the proposed Inf-590K dataset.
As shown in Table~\ref{tab:detection}, our method achieves an mAP of 59.1 and an AP$_{50}$ of 89.7 when using Mask R-CNN as the detection head, surpassing InfMAE (57.1/87.9) by 2.0 and 1.8 points, respectively. Moreover, it consistently outperforms all other baselines, demonstrating the superiority of the DuGI-MAE framework on the infrared object detection task.
 
\subsection{Infrared Semantic Segmentation}
\textbf{Experimental Settings.} 
Experiments on infrared semantic segmentation are conducted using the MSRS dataset~\cite{Tang2022PIAFusion}, which comprises 1,444 pairs of co-registered infrared and visible images. The training set contains 1,083 image pairs, while the test set consists of 361 pairs. In our experiments, only the infrared images are utilized for performance evaluation.
We adopt FCN~\cite{long2015fully} and UperNet~\cite{xiao2018unified} as segmentation heads, and integrate them with the pre-trained DuGI-MAE encoder. The entire model is then fine-tuned in a supervised manner to adapt to the semantic segmentation task.

\textbf{Results and Analyses.} 
The comparison methods including typical semantic segmentation models such as DeepLabV3+~\cite{chen2018encoder}, UperNet~\cite{xiao2018unified}, DNLNet~\cite{ni2022dnl}, DDRNet~\cite{zhang2021ddrnet}, as well as self-supervised learning (SSL)-based methods including MAE~\cite{he2022masked}, MCMAE~\cite{gao2022convmae}, and InfMAE~\cite{liu2024infmae}. The backbones of the SSL-based methods are all pre-trained on the Inf-590K dataset.
Overall, SSL-based methods consistently outperform conventional semantic segmentation models on the infrared modality. As presented in Table~\ref{tab:msrs_segmentation}, integrating the encoders of infrared foundation models, such as InfMAE and DuGI-MAE, with segmentation heads like FCN or UperNet leads to substantial improvements in segmentation performance. These results highlight the effectiveness of self-supervised representations learned from large-scale infrared data. Among all compared methods, DuGI-MAE achieves the highest overall performance, proving its effectiveness and superiority in infrared semantic segmentation.

\subsection{Infrared Small Target Detection}
\textbf{Experimental Settings.} 
Infrared small target detection presents unique challenges due to the inherently limited features of the targets, thereby placing higher demands on the model's ability to learn discriminative representations. To evaluate the performance in this context, we conduct experiments on the IRSTD-1K dataset~\cite{zhang2022isnet}, which comprises 1,000 infrared images with pixel-level annotations. Specifically, we replace the original encoder of IRSTD~\cite{liu2023infrared} with the DuGI-MAE encoder to assess its effectiveness in capturing and representing small target features in infrared imagery.

\textbf{Results and Analyses.} 
We conduct a comprehensive comparison of the proposed method with traditional handcrafted approaches (MPCM~\cite{wei2016multiscale}, IPI~\cite{gao2013infrared}, RIPT~\cite{dai2017reweighted}), fully supervised methods (ACMNet~\cite{qu2021acm}, DNANet~\cite{li2022dense}, UIUNet~\cite{wu2022uiu}, SCAFNet~\cite{zhang2024scafnet}), and self-supervised learning (SSL)-based methods (MAE~\cite{he2022masked}, MCMAE~\cite{gao2022convmae}, InfMAE~\cite{liu2024infmae}) on the IRSTD-1K dataset. For a fair comparison, all SSL-based models adopt the ViT-B backbone pre-trained on Inf-590K and are integrated into IRSTD~\cite{liu2023infrared}.
As demonstrated in Table~\ref{tab:irstd_detection}, the proposed DuGI-MAE achieves the superior performance with an mIoU of 67.1, outperforming MCMAE and InfMAE by 2.8 and 0.6, respectively.
This performance gain can be attributed to the dual-domain mechanism designed for infrared small targets, which effectively enhances the discriminability of target representations by deeply correlating spatial local features with frequency-domain global characteristics.

\subsection{Ablation Study}
\begin{table}[t]
\centering
\begin{tabular}{l c c c}  
\toprule
Method      & Pre-training Data & {mAP} & {AP$_{50}$} \\
\midrule
\multirow{2}{*}{MAE}    & Inf30  & 51.2 & 82.9 \\  
                        & Inf-590K  &  \textbf{52.2} & \textbf{83.8} \\
\midrule
\multirow{2}{*}{MCMAE}  & Inf30  & 56.2 & 87.0 \\
                        & Inf-590K  & \textbf{56.8} & \textbf{87.2} \\
\midrule
\multirow{2}{*}{InfMAE} & Inf30  & 56.5 & \textbf{88.2} \\
                        & Inf-590K &  \textbf{57.1} & 87.9 \\
\midrule
\multirow{2}{*}{DuGI-MAE(Ours)}  & Inf30  & 57.4 & 88.6 \\
                        & Inf-590K  & \textbf{59.1} & \textbf{89.7} \\
\bottomrule
\end{tabular}
\caption{Comparison of self-supervised methods using different pre-training datasets on infrared object detection tasks under the M$^3$FD-inf dataset (ViT-B backbone \& Mask R-CNN detection head).}
\label{tab:infrared_det_results}
\end{table}

\begin{table}[t]
\centering
\begin{tabular}{c c c}  
\toprule
Masking Strategy & {mAP}           & {AP$_{50}$}          \\ 
\midrule
Random mask~\cite{he2022masked}      & 57.2            & 88.1            \\
Gray-values mask~\cite{liu2024infmae} & 58.4            & 88.9            \\
Ours             & \textbf{59.1}   & \textbf{89.7}   \\ 
\bottomrule
\end{tabular}
\caption{Ablation study on masking strategies for DuGI-MAE (ViT-B backbone \& Mask R-CNN detection head).}
\label{tab:masking_ablation}
\end{table}

\begin{table}[t]
\centering
\footnotesize
\setlength{\tabcolsep}{4pt}
\begin{tabular}{l c c S[table-format=2.1] S[table-format=2.1]}
\toprule
Method & Backbone & Model & {mAP} & {AP$_{50}$} \\
\midrule
MAE & ViT-B & Mask R-CNN & 52.2 & 83.8 \\
MAE+DDG & ViT-B & Mask R-CNN & \textbf{53.1} & \textbf{84.9} \\
\midrule
MCMAE & ViT-B & Mask R-CNN & 56.8 & 87.2 \\
MCMAE+DDG & ViT-B & Mask R-CNN & \textbf{57.2} & \textbf{88.1} \\
\midrule
InfMAE & ViT-B & Mask R-CNN & 57.1 & 87.9 \\
InfMAE+DDG & ViT-B & Mask R-CNN & \textbf{57.6} & \textbf{88.4} \\
\bottomrule
\end{tabular}
\caption{Performance improvement of DDG Module on different self-supervised pre-training models.}
\label{tab:ablation_DDG}
\end{table}

\textbf{Comparisons on Pre-trained Models.}
To quantify the impact of pre-training dataset scales, we pre-train MAE, MCMAE, InfMAE and the proposed method on infrared dataset of varying sizes, including Inf30~\cite{liu2024infmae} and the larger Inf-590K. The experimental results on the M$^3$FD-inf dataset are presented in Table~\ref{tab:infrared_det_results}. It can be observed that all self-supervised methods consistently achieve higher mAP when pre-trained on the larger Inf-590K dataset, underscoring the critical role of large-scale infrared pre-training in improving downstream detection performance. 

\textbf{Masking Methods.}
We compare three different masking strategies in Table~\ref{tab:masking_ablation}: 1) Random Masking~\cite{he2022masked}, 2) Information-aware Masking~\cite{liu2024infmae}, and 3) Entropy-based Masking (ours). The information-aware masking ranks tokens by grayscale intensity and retains 25\% tokens using a fixed sampling stride of 4. In contrast, our entropy-based masking selects tokens based on local entropy values, deterministically preserving the top 25\% most informative tokens. 
As shown in Table~\ref{tab:masking_ablation}, the entropy-based masking consistently outperforms alternative methods. We attribute the inferior performance of random masking to its indiscriminate nature, which disrupts the spatial continuity of high-entropy regions, often critical in infrared imagery, resulting in fragmented feature representations.
While the information-aware strategy introduces content sensitivity, its fixed sampling pattern may overlook salient high-entropy regions, resulting in incomplete semantic preservation. In contrast, our entropy-based approach explicitly focuses on preserving the most informative regions, thereby facilitating more effective reconstruction and feature learning.

\textbf{DDG Module.} 
The proposed DDG module is compatible with various pre-trained models that follow the typical encoder-decoder architecture. To evaluate its generalizability and effectiveness, we conduct ablation studies by integrating the DDG module as a feature bridging component between the encoder and decoder in several representative self-supervised frameworks, including MAE~\cite{he2022masked}, MCMAE~\cite{gao2022convmae}, and InfMAE~\cite{liu2024infmae}. As shown in Table~\ref{tab:ablation_DDG}, incorporating the DDG module consistently improves both the mAP and AP$_{50}$ metrics across all baseline models. These results validate the effectiveness of the DDG module in enhancing feature representations and improving downstream task performance.

\section{Conclusion}
In this paper, we propose DuGI-MAE, a self-supervised pre-training framework tailored for the infrared modality, which is inherently characterized by low information density. To enable effective representation learning, we first construct Inf-590K, a large-scale infrared dataset comprising 590,700 images. 
To mitigate the risk of masking informative tokens during pre-training, we propose an entropy-based masking strategy that selectively retains tokens with high information content.
Furthermore, to address the reconstruction bottleneck caused by aggressive masking, we design a Dual-Domain Guidance (DDG) module that incorporates both spatial- and frequency-domain cues. This design enhances the model's ability to capture fine-grained local details and global structural patterns simultaneously. Extensive experiments across multiple downstream tasks, including object detection, semantic segmentation, and small target detection, demonstrate that DuGI-MAE consistently outperforms state-of-the-art methods.
In addition, the DDG module an be seamlessly integrated into existing encoder-decoder pre-training frameworks, further improving their performance on infrared data. 
In future work, we will incorporate more physical priors into the pre-training process to advance the development of infrared foundation models. 

\section{Acknowledgements}
This work was supported in part by the National Natural Science Foundation of China (NSFC) under Grant  62476223, 62576282; in part by the National Key Research and Development Program of China under Grant 2024YFF1306501; in part by Innovation Capability Support Program of Shaanxi (Program No. 2024ZC-KJXX-043); in part by the Natural Science Basic Research Program of Shaanxi Province  (2024JC-DXWT-07).

\bibliography{aaai2026}

\end{document}